\documentclass[letterpaper]{article} 
\usepackage{aaai24}  
\usepackage{times}  
\usepackage{helvet}  
\usepackage{courier}  
\usepackage[hyphens]{url}  
\usepackage{graphicx} 
\usepackage{natbib}  
\usepackage{caption} 
\usepackage{xcolor}
\usepackage{algorithm}
\usepackage{algorithmic}

\usepackage{newfloat}
\usepackage{listings}
\usepackage{booktabs}
\usepackage{multirow}

\DeclareCaptionStyle{ruled}{labelfont=normalfont,labelsep=colon,strut=off} 
\floatstyle{ruled}
\newfloat{listing}{tb}{lst}{}
\floatname{listing}{Listing}
\title{LatestEval: Addressing Data Contamination in Language Model Evaluation through Dynamic and Time-Sensitive Test Construction}

\author {
    Yucheng Li\textsuperscript{\rm 1},
    Frank Guerin\textsuperscript{\rm 1},
    Chenghua Lin\textsuperscript{\rm 2}\thanks{Corresponding author}
}
\affiliations {
    \textsuperscript{\rm 1}Department of Computer Science, University of Surrey, UK\\
    \textsuperscript{\rm 2}Department of Computer Science, University of Manchester, UK\\
    \{yucheng.li, f.guerin\}@surey.ac.uk, chenghua.lin@manchester.ac.uk
}

\usepackage{bibentry}

\begin{document}

\maketitle

\begin{abstract}
Data contamination in evaluation is getting increasingly prevalent with the emergence of language models pre-trained on super large, automatically crawled corpora. This problem leads to significant challenges in the accurate assessment of model capabilities and generalisations. In this paper, we propose LatestEval, an automatic method that leverages the most recent texts to create uncontaminated reading comprehension evaluations. LatestEval avoids data contamination by only using texts published within a recent time window, ensuring no overlap with the training corpora of pre-trained language models. We develop the LatestEval automated pipeline to 1) gather the latest texts; 2) identify key information, and 3) construct questions targeting the information while removing the existing answers from the context. This encourages models to infer the answers themselves based on the remaining context, rather than just copy-paste. Our experiments demonstrate that language models exhibit negligible memorisation behaviours on LatestEval as opposed to previous benchmarks, suggesting a significantly reduced risk of data contamination and leading to a more robust evaluation. Data and code are publicly available at: \url{https://github.com/liyucheng09/LatestEval}.
\end{abstract}

\section{Introduction}

Recent years have seen the ubiquity of pretrained language models in natural language processing (NLP) due to their strong performance and generalisation capability. These models are usually pre-trained on super large internet-crawled corpora. However, many widely used benchmarks are also largely constructed from web resources \cite{hendrycks2020measuring}, which are very likely to be unintentionally included in the pretraining stage. This leads to a major emerging issue called \textit{data contamination}.

Recent analysis has revealed that data contamination is widespread in model evaluations \cite{openai2023gpt4,sainz2023nlp}, which greatly undermines the credibility of evaluation results \cite{marie2023,li2023open} and prevents fair comparisons between models \cite{dickson2023}. Moreover, the massive scale of training data makes decontaminating existing benchmarks extremely difficult \cite{kreutzer-etal-2022-quality}. For many closed models, training data is considered as trade secret and thus confidential, eliminating any possibility for the community to address contamination by decontaminating benchmarks.
One potential solution to avoid contaminated evaluation is to create new test data constantly or use human evaluation \cite{liu2023evaluating,jacovi2023stop}, just like how examination for human works. However, this is extremely inefficient and costly, requiring huge human efforts periodically. 

In this paper, we propose LatestEval, an automatic method that leverages the most recent texts to create novel uncontaminated reading comprehension evaluations. Before starting to create the data, we conduct the \textit{first} manual analysis of real-world Human-AI chat data on document comprehension, to identify the most frequently asked types of knowledge and thus determine the scope of LatestEval.
As for data construction, LatestEval, as a reading comprehension benchmark, typically consists of three components: \textit{passage, query, and answer}. Here we start by 1) collecting recently created texts across various time-sensitive sources as the passages. Three different sources are used in our experiments: arXiv, BBC, and GitHub. We then 2) extract key information from these texts using various tools as the answers. At last we 3) construct questions targeting the extracted information. The questions are generated automatically via template-filling or large language models. During the testing stage, we remove the extracted answers from the original context, which encourages models to infer answers through reasoning based on the remaining context rather than short-cutting via copy-paste. Overall, LatestEval avoids data contamination by using the latest materials for testing, ensuring that it does not overlap with models' pretraining data.

Experimental results show that, in contrast to existing reading comprehension benchmarks where language models show significant memorisation, models exhibit negligible memorisation behaviour on LatestEval. Additional human evaluation and performance experiments further demonstrate LatestEval is reliable and effective in benchmarking state-of-the-art language models.

\section{Data Contamination}

\noindent\textbf{What is data contamination?}~~~Data contamination refers to the phenomenon that examples from the evaluation set are also found in the training data. This might lead to the evaluation failing to accurately reflect models' capabilities, as models can cheat by memorising instead of learning to generalise. There are two primary types of data contamination \cite{dodge2021documenting}: (i) \textit{input-only contamination} refers to only the input appearing in the pretraining corpus, and (ii) \textit{input-and-label contamination} is when both inputs and their labels are present. The latter is generally more problematic, as models can directly memorise input-output pairs. However, the former can still cause issues as models may gain an advantage by learning extra information from the context \cite{li2023open}.

\noindent\textbf{How common is data contamination?}~~~Data contamination appears to be quite widespread across commonly used NLP benchmark datasets based on findings from recent studies. \citet{dodge2021documenting} revealed exact match contamination rates ranging from under 2\% to over 50\% on various GLUE benchmarks when compared to the C4 pretraining data. The GPT-3 study \cite{brown2020language} found over 90\% of examples in Quac, SQuADv2, and DROP were flagged as contaminated. FLAN \cite{wei2021finetuned} evaluations identified 7 out of 26 datasets exhibiting a serious contamination ratio of 50\% and over. LLaMA 2 \cite{touvron2023llama} reported over 16\% of MMLU examples are contaminated and about 11\% are seriously contaminated (more than 80\% token leakage). GPT-4 \cite{openai2023gpt4} uses academic exams instead of NLP benchmarks for model evaluation. While 4 out of 34 exams are found to have zero contamination (e.g., Leetcode and Bar Exam), 9 out of 34 showed over 20\% of instances are marked as dirty examples. \citet{li2023open} illustrate varying level of contamination ranging from 2\% to 45\% in MMLU, C-Eval \cite{huang2023ceval}, Hellaswag \cite{zellers2019hellaswag} and other popular QA benchmarks.


\noindent\textbf{How to identify data contamination?}~~~\citet{dodge2021documenting} takes a straightforward approach to detect exact matches between test set examples and the pretraining data after normalising for capitalisation and punctuation. The \textit{exact match here} means the entire input of an evaluation text is found in the training data. The GPT-3 paper \cite{brown2020language} uses n-gram overlap to identify contamination, treating any examples with 13-gram co-occurrence in both test sets and training data as dirty examples. LLaMA-2 matches on tokenized prompts and takes a bottom-up, token-level approach to identify contamination. Overall, existing approaches usually use substring matching between evaluation examples and training data to identify data contamination. However, if we have no access to the training data, which is often the case for most recent closed models, it is extremely difficult to reveal contamination by observing the models themselves. Pioneering studies propose to identify benchmark data contamination by using search engine \cite{li2023open}, measuring perplexity of test examples \cite{li2023estimating}, or asking models to reconstruct test examples verbatim \cite{carlini2021extracting,carlini2022quantifying}.

\noindent\textbf{To what extent does data contamination affect model evaluation?}~~~While contaminated data can potentially inflate scores, models do not necessarily perform worse on clean subsets or better on dirty subsets across all datasets. The degree of impact likely depends on factors like the dataset characteristics, model scale, and nature of the pre-training data. For instance, GPT-3 \cite{brown2020language} showed a small 1-2\% performance drop on clean subsets for PIQA and ReCoRD, compared to a significant 6\% drop on clean set of SQuAD as 94\% of its test examples were contaminated. The LLaMA model \cite{touvron2023llama} did not show significant gaps between clean and dirty subset performance. On HellaSwag, LLaMA's 70B model showed a 15.3 point gap between clean (63.5) and dirty (78.8) subsets. \citet{li2023open} reveal contamination can inflate benchmark score even if the contamination does not give away the answer (i.e., \textit{input-only contamination}). It also find larger models obtain more advantages from data contamination due to their more powerful memorisation.

\section{Benchmarks for Language Models}

Clean and robust benchmarks are the key to guide further progress of various models in NLP. 
Popular benchmarks used to evaluate large language models include:
\begin{itemize}
    \item Comprehensive: MMLU, Big Bench Hard, AGI Eval
    \item Commonsense reasoning: PIQA, SIQA, HellaSwag, WinoGrande, ARC, OpenBookQA, CommonsenseQA
    \item World knowledge: NaturalQuestions, TriviaQA
    \item Reading comprehension: SQuAD, QuAC, BoolQ
    \item Math: GSM8K, MATH
    \item Code: HumanEval, MBPP
\end{itemize}
where most of them are collected from freely available online sources, which makes them very susceptible to data contamination. For instance, BoolQ \cite{clark2019boolq} heavily relies on Wikipedia articles in their instances construct, which leads to a significant contamination rate of 60\% as shown in \cite{brown2020language} because Wikipedia serves as a key part of GPT-3's pretraining data.

\citet{jacovi2023stop} propose three strategies to alleviate data contamination issues including encrypting test data, refusing derivative distribution, avoiding using internet data that appears with its solution, etc. Our method introduces a novel possibility motivated by examination for humans, where new tests are created dynamically to avoid cheating. Via updating periodically, LatestEval ensures to use of only the most recent texts on the web that were created after any training data were constructed, inherently mitigating the contamination risks.

\section{The Scope of LatestEval}\label{scope}

\begin{table*}[t]
    \centering
    \begin{tabular}{p{4cm}rp{10cm}}
    \toprule
       Query Type & Frequency & ~~~~~~~~~~~~~~~~~~~~~~~~~~~~~~~~~~~~~~~~~~~~~~~~Examples \\
    \midrule
        \multirow{4}{4cm}{Asking for explanation, details or comparison} & 471 & Can you provide more information on how GANs jointly optimize both privacy and utility? \\
        && Do both approaches improve with more training data? \\
        && What is the Accelerated Failure Time (AFT) model? \\
        \midrule
        \multirow{2}{4cm}{Asking for summary, or key insight} & 272 & What is this paper about?\\
        && What is the main conclusion of the analysis section? \\
        \midrule
        \multirow{4}{4cm}{Asking for reason, purpose or benefit} & 144 & What is the purpose of the two-level model? \\
        && What is the role of landmarks in the proposed method? \\
        && How does RANSAC contribute to the multispectral photometric stereo method? \\
        \midrule
        \multirow{2}{4cm}{Asking for examples or demonstrations} & 48 & Can you give an example in terms of how action-value functions works \\
        && Can you explain what is Async-cloud update with an example? \\
        \midrule
        \multirow{2}{4cm}{Asking for future prediction} & 40 & What are the potential application of the MIPS method? \\
        && How could highway connection be used in further architecture design? \\
        \midrule
        \multirow{2}{4cm}{Asking for whether something is presented} & 21 & Did the authors mentioned any rule-based method? \\
        && Have the authors included SeqGAN in their experiments? \\ 
    \bottomrule
    \end{tabular}
    \caption{Six most frequent categories of user query to comprehend papers.}
    \label{tab:scope}
\end{table*}

To determine the scope of LatestEval and make it a meaningful and realistic benchmark, we conduct the first manual analysis on \textit{real-world conversations} between users and AI assistants chatting about documents. This analysis is to reveal which types of information humans find most essential when comprehending texts. We greatly thank \texttt{paperpersichat.tech} for providing real world conversation data about user interaction with chatbots aiming to understand academic papers. We sample 1000 real user queries and manually categorise their intention. We identified six major query categories as illustrated in Table \ref{tab:scope}.

As shown in the table, queries requesting explanations or details are dominant, highlighting the importance of models' capability to locate definitions and elaborate key information to aid user comprehension. Summary requests were also very prevalent, suggesting summarising skills are also critical for reading comprehension. There are also queries focusing on purpose analysis, giving examples, making future predictions, and judging whether something is presented. Note that although this finding is based on conversations about papers, we believe these categories also generalise to other sources in LatestEval i.e., news articles and GitHub readme documents. As a result, we focus LatestEval on the most frequent \textit{five} categories of the query to align with practical needs: 1) terminology explanations and comparison; 2) summarisation; 3) finding the purpose; 4) providing examples; 5) predicting about the future. As a result, LatestEval will extract these corresponding types of key information to allow construction questions that assess models' abilities on these aspects.

\section{LatestEval Pipeline}

\begin{figure*}
    \centering
    \includegraphics[width=\textwidth]{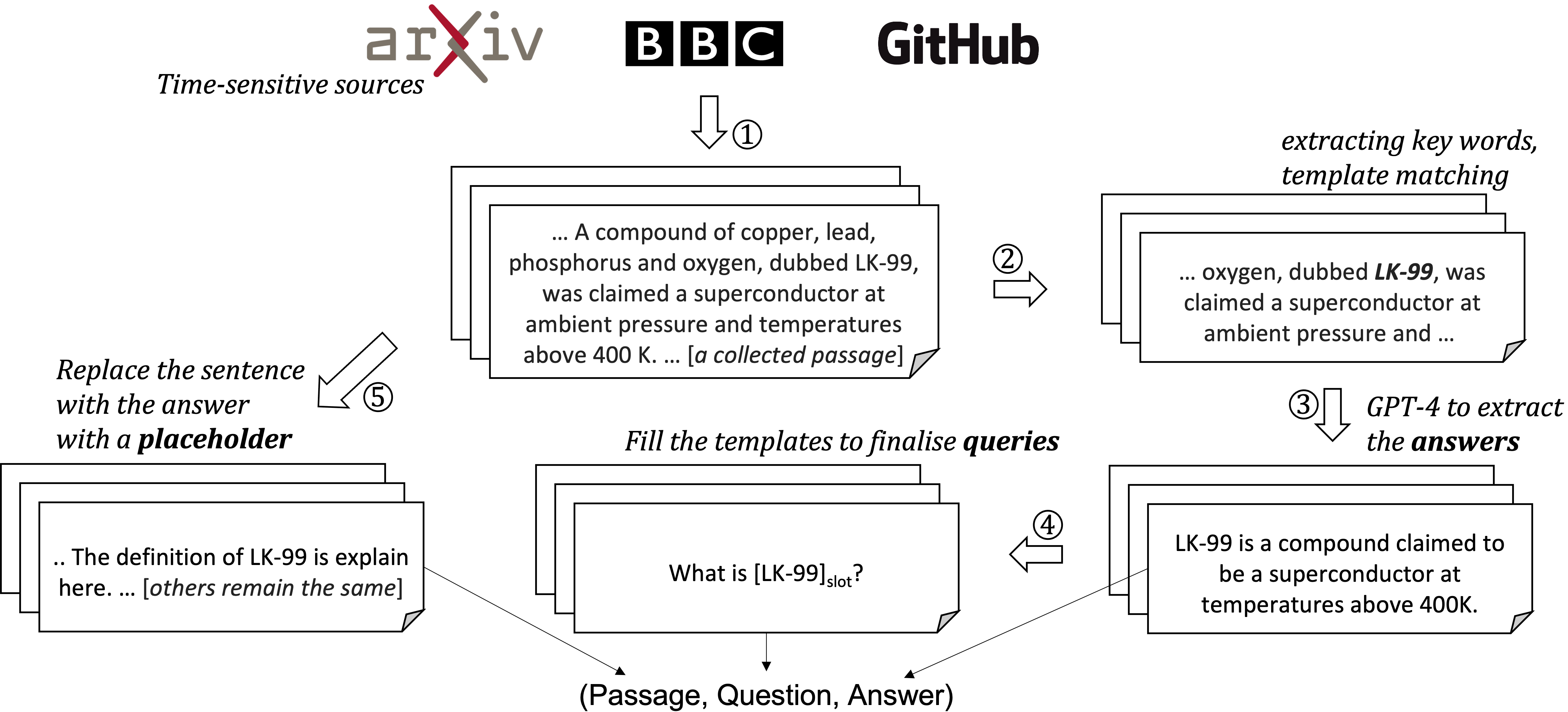}
    \caption{The overall pipeline of LatestEval. Step \textsc{1} is for collecting the latest texts; \textsc{2,3} are to construct the answers; \textsc{4} is to construct corresponding queries; and \textsc{5} is to prepare the passages.}
    \label{fig:pipeline}
\end{figure*}

LatestEval is a \textit{dynamic} reading comprehension benchmark, similar to SQuAD \cite{rajpurkar2018know} and BoolQ, where each test instance is composed of three essential components: a passage, a query, and an answer. Models are required to answer the query based on the passage given. The LatestEval involves a three-stage pipeline: (i) the collection of the most recent texts from the web as the passages, (ii) the extraction of key information as answers, and (iii) the generation of questions corresponding to answers.

\subsection{Collecting Latest Created Texts}

We collect our passages from three different sources: arXiv, BBC, and GitHub. These three sources are continually updated providing real-time created content, which can effectively avoid data contamination in evaluation. In addition, they cover a wide range of topics and areas, offering both formal and informal language usage. This allows a diversified evaluation of model capabilities. At last, their popularity can ensure a sufficient volume of fresh content within even a limited time window.

To collect texts from arXiv, we request latex source of papers via its web API\footnote{http://export.arxiv.org/api/query}, followed by latex parser \texttt{pylatexenc} to extract plain text from \texttt{.tex} files. We collect news articles that appear on the front page of bbc.com to make sure they are fresh. GitHub repositories are open-source programming project, which normally has a \texttt{readme.md} file as a manual. We take the \texttt{readme} files as the passages of read comprehension tests. We use GitHub API\footnote{https://api.github.com/search/repositories} to obtain the latest created repositories and their \texttt{readme} files, but filter out repositories with no or very short (less than 100 words) readme files.

Considering the context window of current language models, we choose to set 1800 words as the max length of LatestEval instances. Therefore, we use only the first section (usually introduction) for arXiv papers, and filter out texts collected from BBC and GitHub that are longer than 1800 words. For a time span of 7 days from 01/07/2023 to 08/07/2023, 
we have collected 2,150 valid papers from arXiv, 525 articles from BBC, and 951 \texttt{readme} files from GitHub that meet our requirements. This demonstrates that there is sufficient content for the next steps. 

\subsection{Answer Construction}

To construct meaningful questions for reading comprehension, we need to identify key pieces of information from the collected passages that can serve as the answers. Based on the scope analysis above, we focus on extracting the five major types of key information.

\noindent\textbf{Terminology explanations.} We employ a two-step strategy to extract terminology explanations and comparisons from the input passage: keyword extraction + large language models. We first process the input passage via the KeyBERT \cite{grootendorst2020keybert} algorithm to extract candidate keywords and phrases. For passages exceeding 512 tokens (max length of BERT), we use \texttt{gpt-3.5-turbo}\footnote{https://platform.openai.com/examples/default-keywords} to extract key terms. Then, we take the key terms and the first few sentences containing the terms as input, and ask GPT-4\footnote{GPT-4 is a closed source model, you can access it via OpenAI API. Check https://platform.openai.com/docs/guides/gpt} to identify their definition from these sentences. Here we only consider terms with an explicit definition in the passage, then we ask GPT-4 to extract the definition as the target answers.

\noindent\textbf{Summarisation.} We use template-matching to directly identify summative content from passages. For papers, we extract summative content indicated with the \textit{abstract} latex environment, key findings at the end of the introduction section, and sentences start with \textit{In summary}. For news, we use \textit{Tl;dr} as the indicator or take the title plus article description (written by editors) as the summary. For GitHub repositories, we take the repository description plus the first few lines of the readme file as the summary of the project.

\noindent\textbf{Others.} We use a pipeline consisting of phrase-matching plus GPT-4 post-processing to construct answers for purpose, examples, and future prediction types of queries. We first use phrase matching to locate these answers in the given passage. For instance, we locate answers for purpose types of queries by retrieving sentences containing the following phrases: \textit{because}, \textit{aim to}, \textit{allow .. to }, \textit{contribute to}, \textit{lead to}, and \textit{motivation}. For examples types of queries, we choose phrases of \textit{for example}, \textit{e.g.,}, \textit{such as}. Note that we notice examples in GitHub readme files are usually involving code generation. We do not consider providing examples type of query for the source of Git Hub. For queries about making future predictions, we use phrases of \textit{future works}, \textit{Forecasts show}, \textit{upcoming features}. Then we ask GPT-4 to extract the analysis about purposes as the target answers.

After extracting answers from the passages, we replace sentences containing these answers with placeholder sentences. For instance, we replace the sentence \textit{``We chose to look at loss because it tends to be less noisy than other measures.''} with \textit{``The reason to look at loss is explained here.''} Here the placeholder sentence is obtained by GPT-4 and template filling. We do this replacement to encourage models to infer answers instead of copy-past.

\subsection{Query Construction}

We propose a hybrid approach with templates and GPT-4 to construct natural queries targeting the extracted answers in the previous steps.
For the summarisation queries, we directly apply simple templates like ``\textit{what is this passage about?}", ``\textit{what are the main points raised in this article?}", which already do a good job in providing clear questions asking for analysis main points. For other question types, we generate questions by providing GPT-4 with the extracted answers and template candidates and asking it to produce questions by filling the templates with appropriate terms. First, we manually create 3-5 templates for each query type. We then feed GPT-4 the answer text along with the template options. GPT-4 selects the most suitable template and fills it with appropriate terms or phrases from the answer, generating a custom question targeting the specific information. 

\begin{figure*}[t]
    \centering
    \includegraphics[width=0.8\textwidth]{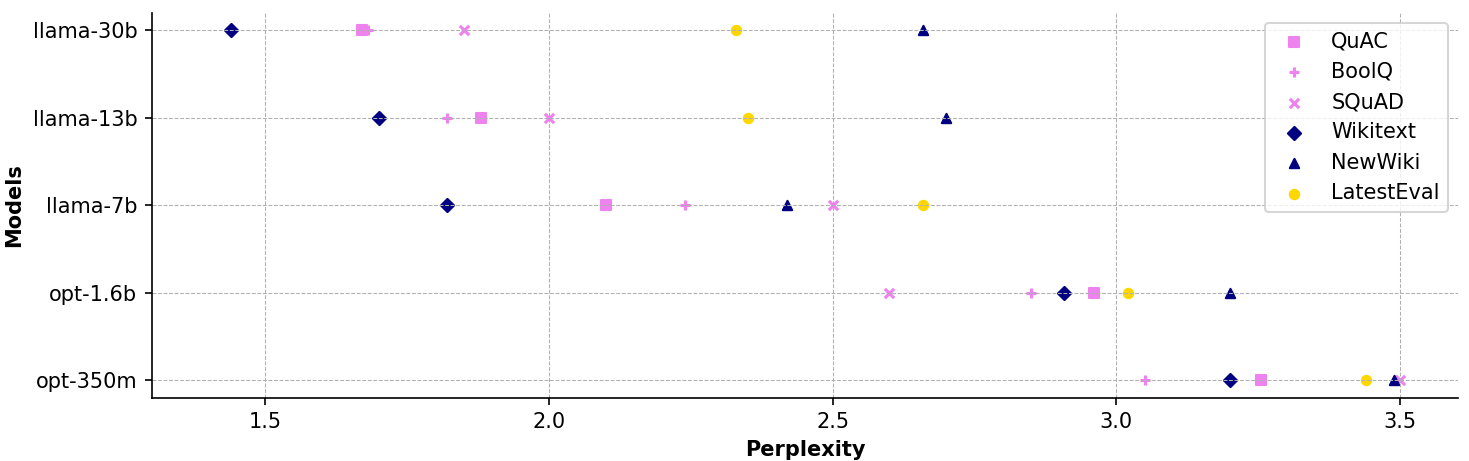}
    \caption{The comparison of datasets' perplexities indicates the contamination extent on various language models.}
    \label{fig:contamination}
\end{figure*}

\subsection{Examples and Statistics}

\begin{table}[]
    \centering
    \begin{tabular}{lrrr|r}
    \toprule
         & Arxiv & BBC & Github & ALL  \\
    \midrule
    Terminology & 1792 &  98 &  825 & 2715 \\
    Summary     &  570 & 504 &  466 & 1540 \\
    Purpose     &  793 & 565 &  421 & 1779 \\
    Example     &  421 &  16 &  113 & 550  \\
    Future      &  147 & 105 &   63 & 315  \\
    \midrule
    ALL         & 3723 & 1288 & 1888 & 6899 \\
    \bottomrule    
    \end{tabular}
    \caption{Statistics of LatestEval, July week 1 2023.}
    \label{tab:statistics}
\end{table}

Here we present the statistics of LatestEval on time slice of week 1 July 2023 as we use this version in our experiments section. In addition, we show the prompts we use in the pipeline of LatestEval, and show some of the templates we use in the query construction. As shown in Table \ref{tab:statistics}, we have constructed 6899 reading comprehension tests in total from 500 passages for each of the three sources, and 3723, 1288, and 1888 tests from arXiv, BBC and GitHub, respectively. We do not use all generated data in our experiments, we sample 3k examples based on the distribution of question types we find in the real-world data analysis, i.e., Table \ref{tab:scope}. We have shown an example of how the overall procedure for a test case construction in Figure \ref{fig:pipeline}. We explain the prompt we use in steps 3 and 5, as GPT-4 is heavily used in these two steps. In step 3, we use the prompt: \textit{``\{relevant\_sentences\}. Please extract the definition of the term \{term\} that the above sentences explicitly defined. You must just do copy-pasting and be faithful to the given passage."}. The prompt in step 5 is \textit{``\{the\_query\}, \{relevant\_sentence\}. Based on the above given information, generate a placeholder sentence considering the following examples \{placeholder\_examples\}."}

\section{Experiments}

We performed three experiments to evaluate our LatestEval: contamination test, performance test, and human evaluation.

\subsection{Contamination Test}
First, we conduct a contamination test on common reading comprehension benchmarks to measure the extent that models memorise their test instances during pretraining. Although most existing analyses identify contamination by computing the overlapping between training and test sets, 
we cannot use the same approach as the pretraining data of current language models are mostly not publicly available.
Therefore, we propose a novel method to quantify test contamination by observing whether models exhibit memorisation behaviour on test instances, which can serve as a strong signal of test contamination. \citet{carlini2021extracting,carlini2022quantifying} have defined the ``memorisation" of models that a sequence is considered as memorised if the model has considerably smaller perplexity on that sequence. The idea is that sequences leak in the training data will tend to have lower perplexity (i.e., higher likelihood) than sequence models never seen before. This is well demonstrated in Figure \ref{fig:contamination}, where we compare the perplexity of two Wikipedia subsets: the first is \texttt{Wikitext} \cite{merity2016pointer} which is widely used in language model pretraining, the second is \texttt{NewWiki}, latest Wikipedia texts that created after April 2023, after all tested models were released. We find that Wikipedia content learned during pretraining has a much lower perplexity than recently created Wikipedia content.

\begin{figure*}[t]
    \centering
    \includegraphics[width=\textwidth]{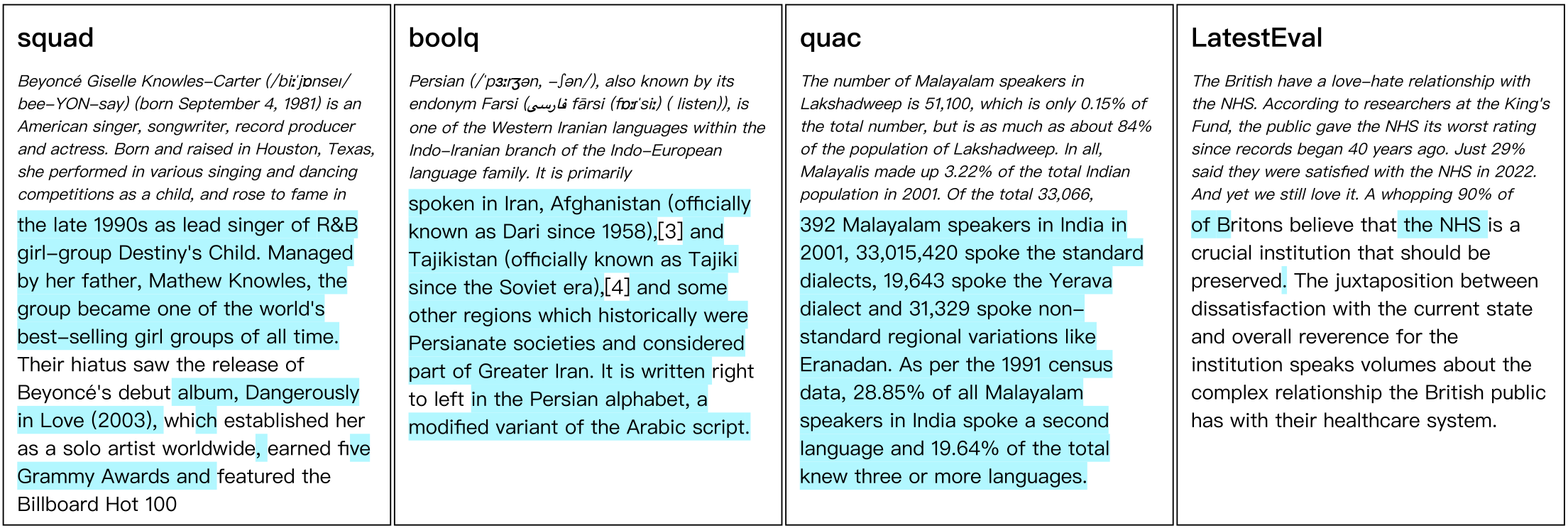}
    \caption{Memorisation test of \texttt{GPT-4} model on four benchmarks. Coloured text refers to the text generated by \texttt{GPT-4} that matches the original test text. The four examples shown are just the first instance of each benchmark, so no cherry picking.}
    \label{fig:visualisation}
\end{figure*}

Now we compute the perplexity of the validation set on the three existing reading comprehension benchmarks SQuADv2 \cite{rajpurkar2018know}, BoolQ \cite{clark2019boolq}, QuAC \cite{choi2018quac} and compare them to LatestEval. We involve five foundation models here: \texttt{opt-350m}, \texttt{opt-1.3b} \cite{zhang2022opt}, \texttt{Llama-7b}, \texttt{Llama-13b}, and \texttt{Llama-30b} \cite{touvron2023llama}.
OPT models are run with fp32 and others are run with fp16. We make sure the input texts of each benchmark have the same length so that we can obtain a fair comparison.

\noindent\textbf{Results.} As shown in Figure \ref{fig:contamination}, the perplexities of BoolQ and QuAC are very close to \texttt{wikitext} and are significantly lower than freshly created Wiki content across all language models. This reflects clear memorisation of language models on test examples in these two benchmarks, and thus can be considered as a signal of data contamination. The perplexities of SQuADv2 are divergent, which has a rather high perplexity and is close to freshly created Wiki texts on small models such as \texttt{llama-7b} and \texttt{opt}s. The perplexity decreases significantly on larger models such as \texttt{llama-13b,30b}. Based on this observation, we believe test examples in SQuADv2 were not memorised by small models but the extent of being contaminated increases with model scale. At last, we find a clear trend that LatestEval constantly has a perplexity very close to or even higher than fresh Wiki text, which shows that models have no prior knowledge of our constructed benchmark and thus have considerably less contamination risk.

In Figure~\ref{fig:visualisation}, we also show how \texttt{GPT-4} memorise benchmark test examples by real cases. We prompt the model with a prefix and check whether they are able to reproduce the test examples verbatim, following the setting in \cite{carlini2022quantifying}. We find BoolQ and QuAC are heavily contaminated, followed by SQuADv2 that is also partly affected. LatestEval tends to not be affected by the contamination issue. The results here well align with our finding in the perplexity-based analysis (Figure \ref{fig:contamination}).

\subsection{Performance Test}

We test some leading large language models using our LatestEval. We include both foundation models as well as language models after supervised fine-tuning and human feedback reinforcement learning, including \texttt{GPT-3.5-turbo}\footnote{also known as ChatGPT}, \texttt{GPT-4}, \texttt{llama-13b}, \texttt{llama-30b}, and \texttt{vicuna-13b} \cite{vicuna2023}. We are aware that the \texttt{llama-2} series has been recently released. However, our experiments were based on the July week 1 version of LatestEval. Given what \texttt{llama-2} has reported that their models incorporated data from July 2023 in their training, including \texttt{llama-2} would not ensure a fair comparison since they might have been exposed to passages from LatestEval. We will definitely add the \texttt{llama-2} series in future evaluations.

Instead of relying on $n$-gram based metrics such as BLEU and ROUGE to compare models' answers to reference answers, we utilise a LLM-as-a-judge method \cite{zheng2023judging} which has proven to be effective and robust for evaluating models' generations. LLM-as-a-judge provides two grading systems, where the first is single-answer grading and the second is pair-wise win rate. We report the results of both and visualise the results with their software. We adopt the prompt template from InstructGPT \cite{ouyang2022training} designed for reading comprehension test and test models on a zero-shot basis. 
Due to the monthly limit of \texttt{GPT-4} quota, we do not use all pair-wise data, instead, we construct 6k pair-wise data to calculate the pair-wise win rate.

\begin{figure*}
    \centering
    \includegraphics[width=0.9\textwidth]{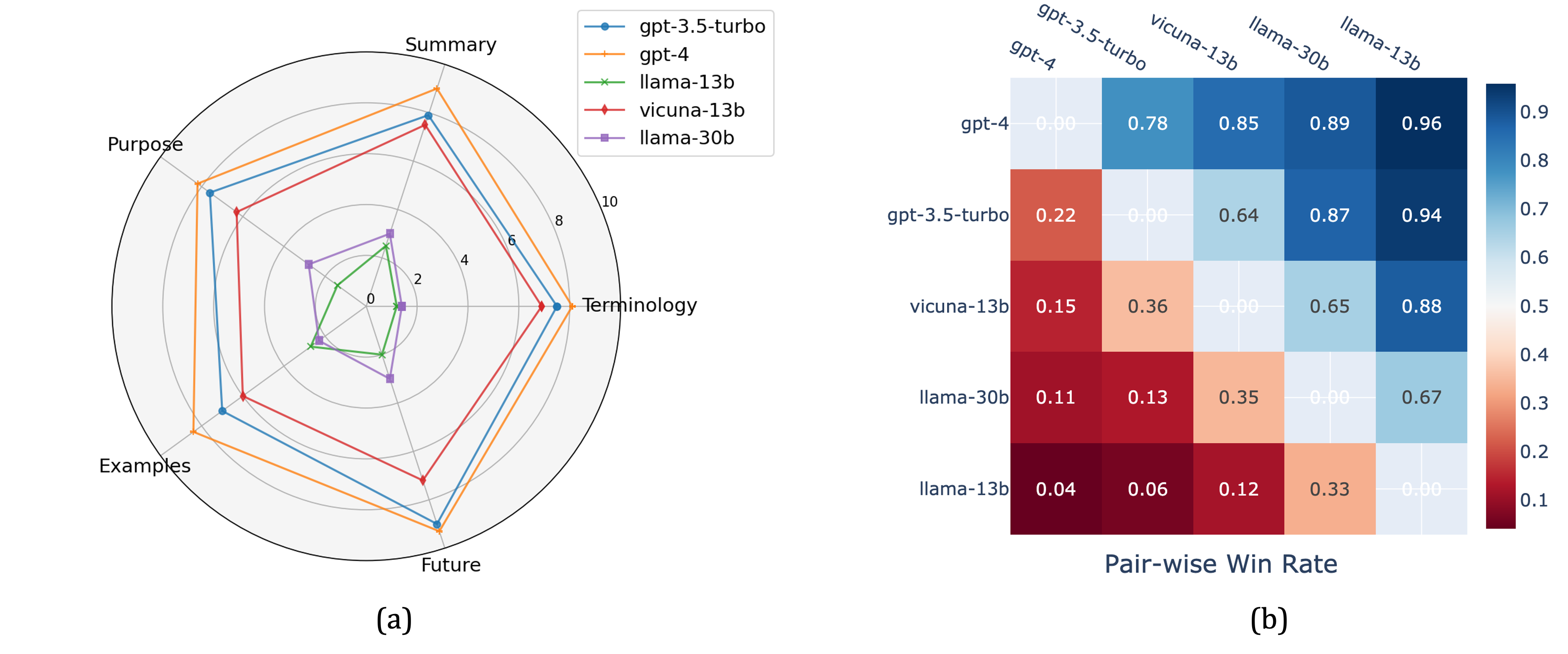}
    \caption{\texttt{(a)}: single answer scores across five types of queries; \texttt{(b)}: pair-wise win rate, y-axis indicates the winner. }
    \label{fig:performance}
\end{figure*}

As shown in Figure \ref{fig:performance}, LatestEval is effective in differentiating large language models. The pair-wise win rate show that \texttt{GPT-4}, and \texttt{gpt-3.5-turbo} beat other models significantly. 
Foundation models without further tuning such as \texttt{llama-13b} and \texttt{llama-30b} hardly generate reasonable outputs, and thus perform rather poorly. Compared to previous reading comprehension benchmarks, LatestEval enables a fine-grained analysis on five categories of questions. \texttt{GPT-4} demonstrates better performance across all types of queries. \texttt{Vicuna} is good at summary and terminology explanation, but less competitive on providing future predictions and examples. In addition, we find models struggle most with terminology explanation and identifying purpose, which indicates room for future improvement.

\subsection{Human Evaluation}
To further validate the robustness of LatestEval, we conducted human evaluations.
LatestEval, due to its automatic construction nature, can potentially lead to three issues: 1) as the use of GPT-4 for extracting answers and filling query templates, we shall evaluate how faithful GPT-4 can be to the given passage and measure how reliable are the extracted answers; 2) to encourage reasoning instead of copy-pasting, we delete the explicit answers from the original passage, which might lead to unanswerable questions; 3) even we have deleted the existing answers, the answers might be mentioned multiples times in the passage, so we also evaluate whether models can cheat by finding the answers explicitly somewhere else in the input after the answer deletion. 
Our human evaluation is to measure and analyse the above three aspects. We sample 200 documents from LatestEval July week 1 and analyse their corresponding 783 reading comprehension pairs. Five annotators from Amazon Turk are employed and each test example is annotated at least three times on three dimensions: \textit{faithfulness}, \textit{answerability}, and \textit{copyability}.

\begin{table}
    \centering
    \small
    \begin{tabular}{lrrrr}
    \toprule
        &  Num. & unfaithful & unanswerable & copyable \\
    \midrule
       Arxiv & 402 & 12 & 22 & 14 \\
       BBC & 220 & 7 & 13 & 2 \\
       Github & 161 & 17 & 11 & 1 \\
    \midrule
    Term. & 309 & 18 & 27 & 6 \\
    Summary & 190 & 0 & 0 & 5 \\
    Purpose & 187 & 13 & 19 & 6 \\
    Example & 62 & 3 & 0 & 0 \\
    Future & 35 & 2 & 0 & 0 \\
    
    \bottomrule
    \end{tabular}
    \caption{Human evaluation on the faithfulness, answerability, and copyability of LatestEval.}
    \label{tab:human_eval}
\end{table}

\noindent\textbf{Results.} We show our human evaluation results in Table \ref{tab:human_eval}, where we analyse the number of unfaithful, unanswerable, and copyable cases for each source and each category of question-answer pair. First, we have identified the faithful issues among all three sources, affecting 4.5\% of examples in total. The issue of faithfulness is mainly from to the answer construction process, which usually happens when GPT-4 is unintended to use its inner knowledge instead of the given context to extract the answers. For example, in \textit{``It has many obvious applications for outdoor scene understanding, from city mapping to forest management."}, the authors explain the potential application of the techniques. But GPT-4 rephrases or adds new things based on its inner knowledge, which we cannot ensure correctness and thus are not expected: \textit{"... potential applications include urban planning, autonomous vehicles, environmental protection, etc."} From the view of query categories, faithfulness issues are mainly from Terminology explanation and Purpose analysis questions.
To avoid model solving questions by just copy-pasting, we delete the explicit answers and ask models to infer the answer from the context. But this approach can somehow lead to unanswerable questions. We discovered unanswerable questions among all sources and two types of questions that affect 5.7\% of questions. However, an interesting finding is that large language models can guess the correct answers even the questions are flagged as unanswerable, e.g. guessing the meaning of a term even the given content has no relevance to the term at all. At last, we analyse whether the model can search and copy the answer from somewhere in the given context, which prevents the reasoning of models. We have detected 2.1\% of examples affected by the copy issue, in terminology, summary, and purpose types of queries.

Based on the above analysis, we believe LatestEval could benefit from 1) a post-processing stage with quality assessment; and 2) use few-shot setting in data construction instead of the current zero-shot setting. By involving a quality assessment procedure, LatestEval could conduct a self-review procedure for each example and then filter out uncertain cases. We could easily customise the quality assessment procedure to further adjust the benchmark. In addition, including demonstrations in the input prompt might enhance the overall robustness of data construction, which can be useful in mitigating the above-mentioned issues. We leave them for future study.

\section{Conclusion}

This paper proposes LatestEval, an automatic test construction pipeline that utilises the latest materials to avoid data contamination in language model evaluation. To determine the scope of LatestEval, we conduct the first analysis of real-world human-AI conversational data to find the most frequently asked queries. We also propose a novel perplexity-based approach to evaluate the extent of benchmark contamination, and find popular language models exhibit negligible memorisation on LatestEval.

\nocite{li2023compressing}
\nocite{li2024evaluating}

\bibliography{aaai24}

\end{document}